\documentclass[conference]{IEEEtran}
\IEEEoverridecommandlockouts

\usepackage{amsmath,graphicx}
\usepackage{xcolor}
\usepackage{booktabs}
\usepackage{multirow}
\usepackage{hhline}
\usepackage{colortbl} 
\usepackage{subcaption}
\usepackage{float}      
\definecolor{C1}{HTML}{93BFCF}
\definecolor{C2}{HTML}{A0C3D2}
\definecolor{C3}{HTML}{BDCDD6}
\definecolor{C4}{HTML}{EEE9DA}
\definecolor{C5}{HTML}{FFF1DC}
\definecolor{C6}{HTML}{E8D5C4}
\definecolor{C7}{HTML}{EEEEEE}
\definecolor{C8}{HTML}{BCEE68}

\usepackage{cite}
\usepackage{amsmath,amssymb,amsfonts}
\usepackage{algorithmic}
\usepackage{graphicx}
\usepackage{textcomp}
\usepackage{xcolor}

\def\BibTeX{{\rm B\kern-.05em{\sc i\kern-.025em b}\kern-.08em
    T\kern-.1667em\lower.7ex\hbox{E}\kern-.125emX}}
\begin{document}

\title{GS-PT: Exploiting 3D Gaussian Splatting for Comprehensive Point Cloud Understanding via Self-supervised Learning
}

\author{
\IEEEauthorblockN{
    Keyi Liu$^{1,\star}$,    
    Yeqi Luo$^{1,\star}$   
    Weidong Yang$^{1}$, 
    Jingyi Xu$^{1}$, 
    Zhijun Li$^{2}$, \IEEEmembership{Fellow, IEEE},  
    Wen-Ming Chen$^{3}$, 
    Ben Fei$^{1}$
    \thanks{$^{\star}$Equal contribution.}
    }
    
    \IEEEauthorblockA{$^{1}$\textit{School of Computer Science, Fudan University}\\$^{2}$\textit{School of Mechanical Engineering, Tongji University}\\$^{3}$\textit{Academy for Engineering and Technology, Fudan University}\\
    {{$\text{23210240242}|\text{23212010018}|\text{jyxu22}|\text{bfei21}$}@m.fudan.edu.cn, wdyang@fudan.edu.cn, zjli@ieee.org, chenwm@fudan.edu.cn}
    }
}  

\maketitle

\begin{abstract}
Self-supervised learning of point cloud aims to leverage unlabeled 3D data to learn meaningful representations without reliance on manual annotations.
However, current approaches face challenges such as limited data diversity and inadequate augmentation for effective feature learning.
To address these challenges, we propose GS-PT, which integrates 3D Gaussian Splatting (3DGS) into point cloud self-supervised learning for the first time.
Our pipeline utilizes transformers as the backbone for self-supervised pre-training and introduces novel contrastive learning tasks through 3DGS.
Specifically, the transformers aim to reconstruct the masked point cloud.
3DGS utilizes multi-view rendered images as input to generate enhanced point cloud distributions and novel view images, facilitating data augmentation and cross-modal contrastive learning.
Additionally, we incorporate features from depth maps.
By optimizing these tasks collectively, our method enriches the tri-modal self-supervised learning process, enabling the model to leverage the correlation across 3D point clouds and 2D images from various modalities. 
We freeze the encoder after pre-training and test the model's performance on multiple downstream tasks.
Experimental results indicate that GS-PT outperforms the off-the-shelf self-supervised learning methods on various downstream tasks including 3D object classification, real-world classifications, and few-shot learning and segmentation.
\end{abstract} 

\begin{IEEEkeywords}
3D Gaussian Splatting, self-supervised learning, pre-training, point clouds, 3D understanding
\end{IEEEkeywords}

\section{Introduction}
\label{sec:intro}


\begin{figure*}[t]
	\centering
	\includegraphics[width=0.8\textwidth]{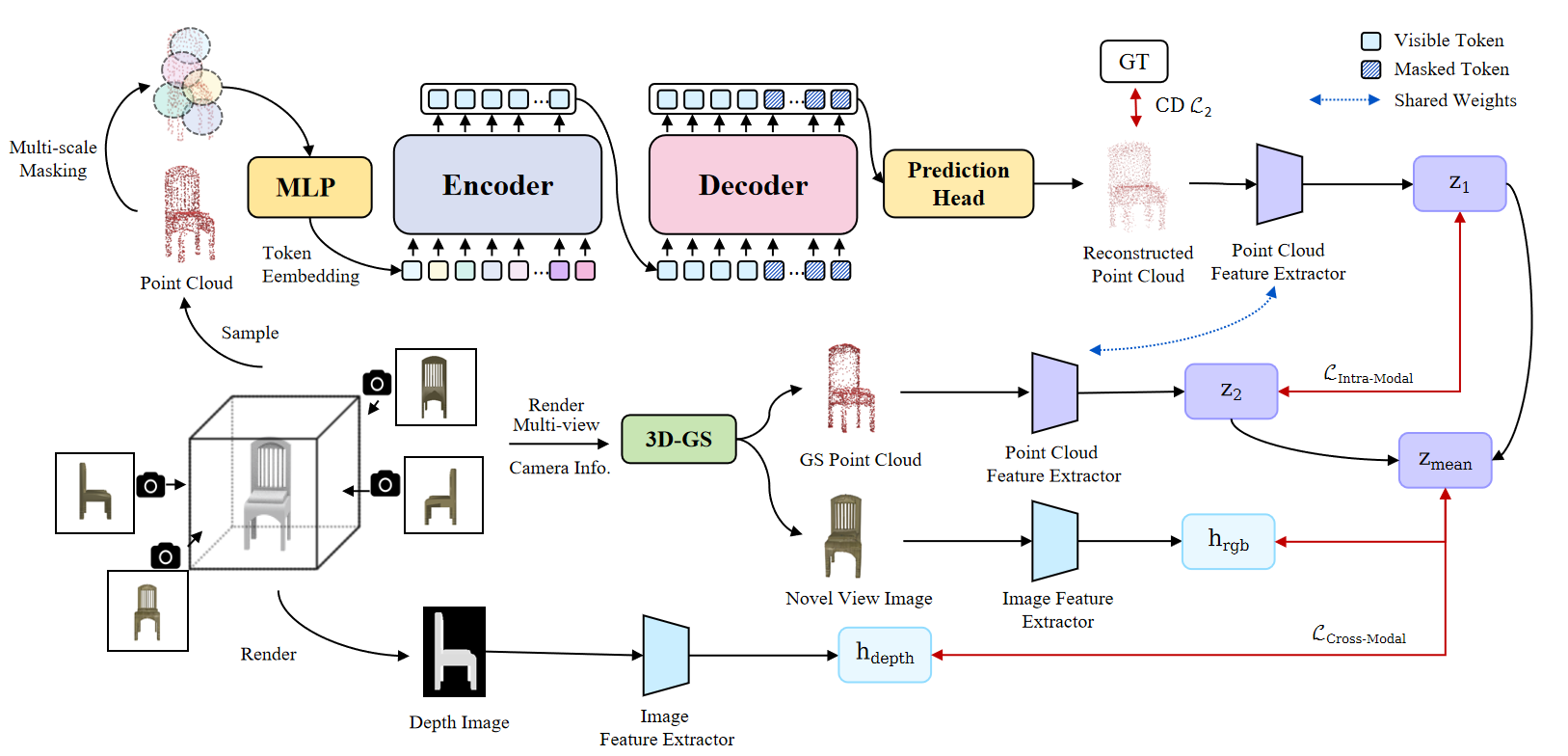} %
    \vspace{-0.3cm}
	\caption{Overview of GS-PT, a unified tri-modal pre-training framework. It is composed of three branches. First, the point cloud is masked, embedded, and fed into the hierarchical encoder-decoder branch, which learns high-level latent features of the point cloud. Second, the 3DGS branch utilizes multi-view rendered images as input, generating enhanced point cloud distributions and novel view images. Additionally, depth maps are rendered. Finally, we create a triplet $T_{\text{i}}$ comprising point cloud $P_{\text{i}}$, novel view image $I_{\text{i}}$, and depth map $D_{\text{i}}$, aligning these modalities into the same feature space using feature extractors.}
	\label{fig:pipeline1}
\vspace{-0.6cm}
\end{figure*}

3D point clouds serve as concise and versatile representations, providing abundant geometric, shape, and scale details, making them a popular choice for 3D data representation. Training of deep neural networks is typically reliant on large-scale annotated datasets. 
However, gathering such annotations of 3D point clouds can be laborious and time-consuming due to challenges like occlusion and irregular structure of point clouds. To mitigate this issue, self-supervised learning stands out as a prominent solution.\par
Self-supervised methods learn visual features from large-scale unlabeled point clouds without relying on any human-annotated labels. 
A popular approach involves designing pretext tasks to train the network by optimizing specific loss functions~\cite{fei2023self}.
However, current paradigms of self-supervised learning (SSL) for point clouds always encounter two main challenges. 
Firstly, effective self-supervised learning requires a comprehensive understanding that integrates information from diverse sources, including point clouds, rendered RGB images, and depth maps. 
However, the high-quality data pairs across these modalities are scarce.
To alleviate it, CrossPoint~\cite{afham2022crosspoint} utilizes rendered RGB images from only 13 object categories for pre-training, which is considerably fewer than the original 55 categories available in the ShapeNet~\cite{chang2015shapenet} dataset. 
Secondly, prevailing discriminative self-supervised learning methods rely on simple geometric transformations for point clouds and images~\cite{afham2022crosspoint,yu2022point,huang2021spatio}.
While such transformations assist in contrastive learning, they often fail to create diverse representations, resulting in a simplistic alignment of features that undermines the model's capacity for robust generalization.\par
Currently, 3DGS has garnered widespread adoption across various domains: surface reconstruction~\cite{guedon2024sugar,chen2023neusg}, dynamic modeling~\cite{yang2024deformable,chen2023periodic}, large-scene modeling~\cite{lin2024vastgaussian}, scene manipulation~\cite{chen2024gaussianeditor,pokhariya2024manus}, generation~\cite{xu2024agg,liang2024luciddreamer}, 3D perception~\cite{zhou2024drivinggaussian} and human modeling~\cite{jiang2024hifi4g,moreau2024human}. 
Leveraging its remarkable ability to synthesize realistic scenes from novel perspectives, point clouds optimized through 3DGS can yield new samples from diverse perspectives, enhancing the model's ability to learn comprehensive geometric features and structural details. Utilizing 3DGS for point cloud self-supervised learning not only augments the training dataset with additional samples but also simulates real-world interferences, consequently bolstering the model's robustness and generalization capability.
\par
To address the two challenges associated with self-supervised learning in point clouds, we propose GS-PT, which leverages 3D Gaussian Splatting~\cite{kerbl20233d} for pre-training a Transformer backbone, enhancing its comprehensive understanding of point clouds.
Firstly, our GS-PT creates scalable multimodal triplets in real-time from the 3D meshes, which include point clouds, RGB images, and depth maps.
We employ a multimodal pre-training pipeline to align these multimodal triplets, thereby enabling the learning of a comprehensive multimodal 3D representation for 3D backbone.
Secondly, unlike existing methods, our data augmentation technique is not confined to simple geometric transformations.
Instead, we integrate 3DGS, which explicitly represents 3D objects and enables novel view synthesis.
By utilizing multi-view rendered images of the original 3D object as input, our method generates enhanced point cloud distributions and novel view images, thus facilitating diverse data augmentation for contrastive learning.

\section{Method}
\label{sec:Method}

\subsection{Training Triplets for GS-PT}
\label{sub:Training_triplets}
The pipeline of our GS-PT is shown in Fig \ref{fig:pipeline1}. We build our dataset of triplets from ShapeNet~\cite{chang2015shapenet}, which contains more than 50,000 CAD models from 55 categories.
For each CAD model $i$ in the dataset, we create a triplet $T_{\text{i}}$ :$(P_{\text{i}}, I_{\text{i}}, D_{\text{i}})$ comprising point cloud $P_{\text{i}}$, novel view image $I_{\text{i}}$ and depth map $D_{\text{i}}$.
The GS-PT will then utilize these triplets for pre-training.

\textbf{Point Cloud Branch.}
We directly use the generated
point cloud of each CAD model in ShapeNet55~\cite{yu2021pointr}.
We uniformly sample $N_{\text{p}} = 2,048$ points from the original point cloud.
Then the hierarchical transformers~\cite{zhang2022point} take the point cloud $P_{\text{i}}$ as input and output reconstructed point cloud $\hat{P}_{i}$.

\textbf{On-the-fly Image Rendering.}
The 3D models in ShapeNet~\cite{chang2015shapenet} do not contain corresponding images.
To get multi-view images of each object, we place virtual cameras around each object and render the corresponding RGB images and depth maps in real-time.
Specifically, we render RGB images from each of the four orthogonal viewing angles and randomly select one of these angles to generate a corresponding depth map. Consequently, for each object $i$, we obtain four RGB images and one depth map $D_{\text{i}}$.
During each pre-training iteration, the RGB images serve as the input for the 3DGS branch, while $D_{\text{i}}$ is utilized as the input for feature extractor $f_{\theta_{D}}(\cdot)$ to extract the depth feature.

\begin{figure}[t]  
    \centering 
    \includegraphics[width=0.9\linewidth]{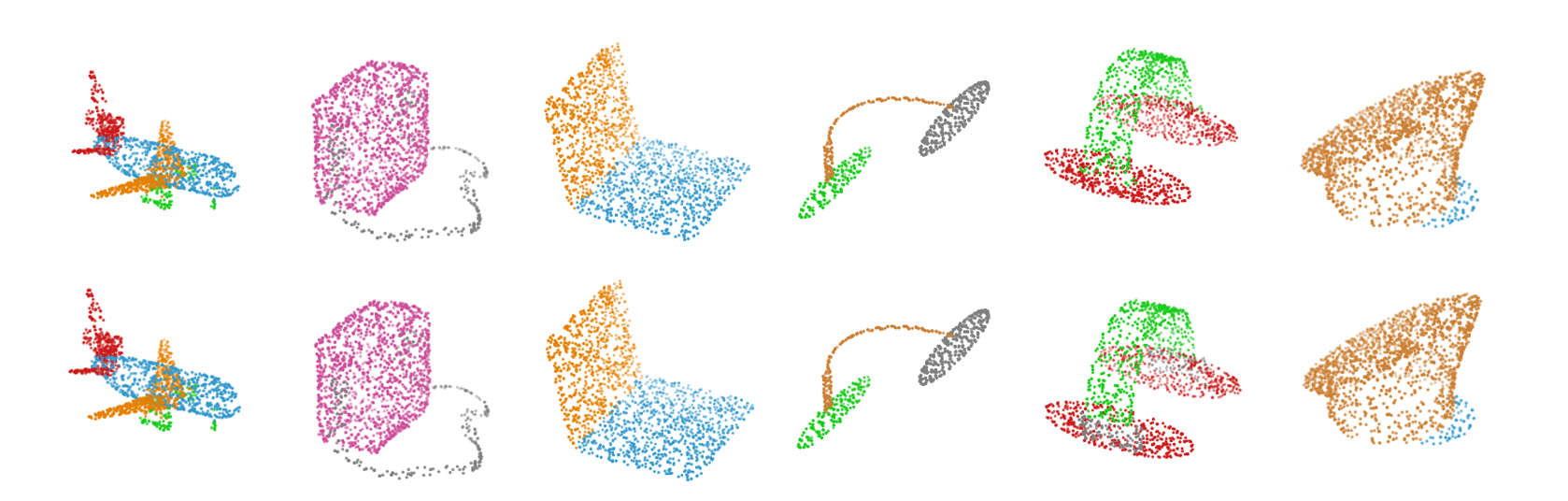} 
    \vspace{-0.3cm}
   \caption{Visualization of part segmentation on ShapeNetPart. The first row is GS-PT, and the second row is ground truth.}  
    \label{fig:Visualization of part segmentation}  
    \vspace{-0.5cm}
\end{figure} 

\begin{figure}[t]  
    \centering  
    \begin{subfigure}[t]{0.32\linewidth} 
        \centering  
        \includegraphics[width=\linewidth]{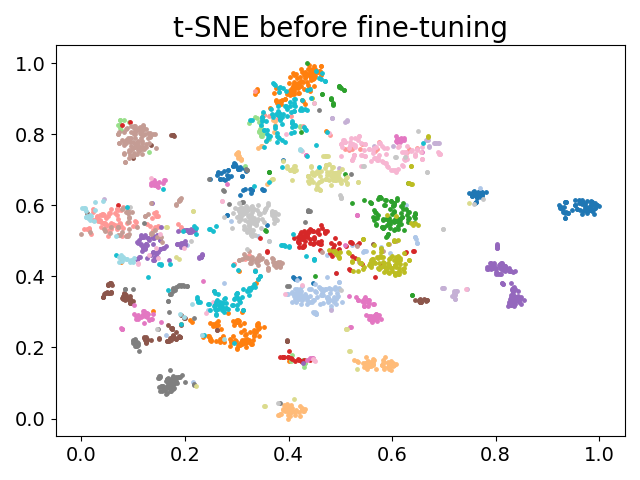} 
        \vspace{-0.8cm}
        \label{fig:tsne_before_modelnetfintune}  
    \end{subfigure}\hfill  
    \begin{subfigure}[t]{0.32\linewidth}  
        \centering  
        \includegraphics[width=\linewidth]{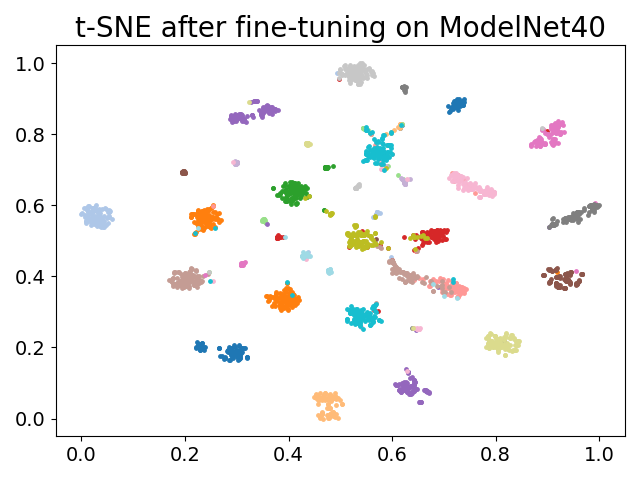}
        \vspace{-0.8cm}
        \label{fig:tsne_after_modelnetfintune_1}  
    \end{subfigure}\hfill  
    \begin{subfigure}[t]{0.32\linewidth}  
        \centering  
        \includegraphics[width=\linewidth]{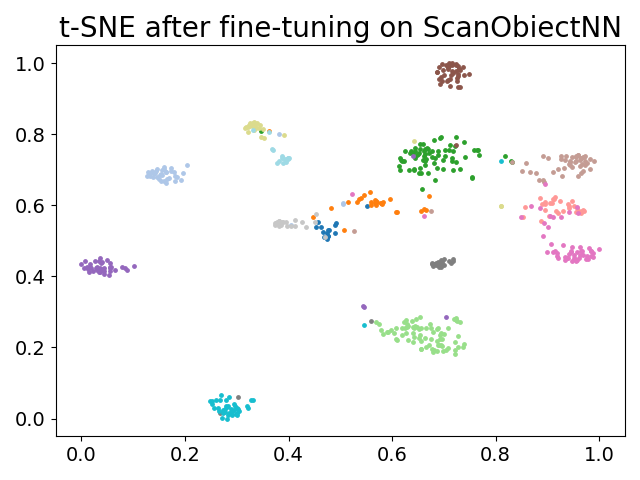} 
        \vspace{-0.8cm}
        \label{fig:tsne_after_ScanObiectNNfintune_2}
    \end{subfigure}  
    \vspace{-0.2cm}
    \caption{Feature visualization using t-SNE. From left to right: Before fine-tuning, fine-tuning on ModelNet40, fine-tuning on ScanObjectNN}  
    \label{fig:tSNE visualization} 
\vspace{-0.8cm}
\end{figure}

\begin{table}[t]
\caption{\textbf{Classification on ModelNet40 dataset.} `Rep.' means we reproduce these methods.}
\vspace{-0.2cm}
\centering
\tabcolsep=0.07cm
\scalebox{0.9}{%
\begin{tabular}{l|l|c}
\hline \toprule \rowcolor{C7!50} 
                                 & Methods                  & Accuracy \\  \midrule
\multirow{10}{*}{Supervised}     & PointNet~\cite{qi2017pointnet}                 & 89.2     \\ \hhline{~|-|-}  
                                 & \cellcolor{C7!50}PointNet++~\cite{qi2017pointnet++}             &\cellcolor{C7!50} 90.3     \\ \hhline{~|-|-} 
                                 & PointWeb~\cite{zhao2019pointweb}                 & 92.3     \\ \hhline{~|-|-} 
                                 & \cellcolor{C7!20}SpiderCNN~\cite{xu2018spidercnn}                  & \cellcolor{C7!20}92.4     \\ \hhline{~|-|-}  
                                 & PointCNN~\cite{li2018pointcnn}                 & 92.5     \\ \hhline{~|-|-} 
                                 & \cellcolor{C7!50}KPConv~\cite{thomas2019kpconv}                   & \cellcolor{C7!50}92.9     \\ \hhline{~|-|-}  
                                 & DGCNN~\cite{wang2019dynamic}                  & 92.9     \\ \hhline{~|-|-}  
                                 & \cellcolor{C7!50}PCT~\cite{guo2021pct}               & \cellcolor{C7!50}93.2     \\ \hhline{~|-|-}  
                                 & PVT~\cite{zhang2022pvt}             & 93.6     \\ \hhline{~|-|-}  
                                 & \cellcolor{C7!50}PointTransformer~\cite{zhao2021point}  & \cellcolor{C7!50}93.7     \\ \hhline{~|-|-}  
                                 & Transformer~\cite{yu2022point}     & 91.4     \\  \midrule
\multirow{8}{*}{Self-supervised} & \cellcolor{C7!50}OcCo~\cite{wang2021unsupervised}                     & \cellcolor{C7!50}93.0     \\ \hhline{~|-|-} 
                                 & STRL~\cite{huang2021spatio}                      & 93.1     \\ \hhline{~|-|-}  
                                 & \cellcolor{C7!50} \begin{tabular}[c]{@{}c@{}}Transformer\\ +OcCo~\cite{wang2021unsupervised} \end{tabular} & \cellcolor{C7!50}92.1     \\ \hhline{~|-|-}    
                                 & Point-BERT~\cite{yu2022point}        & 93.2    \\ 
                                 \hhline{~|-|-} 
                                 & \cellcolor{C7!50}Point-MAE~\cite{pang2022masked}      & \cellcolor{C7!50}\underline{93.8}     \\
                                 \hhline{~|-|-}  
                                 & Point-MAE (Rep.)        & 93.1 \\ 
                                 \hhline{~|-|-} 
                                 & \cellcolor{C7!50}Point-M2AE~\cite{zhang2022point}     & \cellcolor{C7!50}\textbf{94.0}     \\ 
                                 \hhline{~|-|-} 
                                 & Point-M2AE (Rep.)        & 93.5     \\ 
                                 \hhline{~|-|-} 
                                 & \cellcolor{C7!50}\textbf{GS-PT}      & \cellcolor{C7!50}\underline{93.8}     \\
                                 \bottomrule
\end{tabular}%
}
\vspace{-0.4cm}
\label{table:modelnet40}
\end{table}

\begin{table}[t]
\caption{\textbf{Classification on ScanObjectNN.} Accuracy ($\%$) on three settings of ScanObjectNN are listed. `Rep.' means we reproduce these methods.}
\vspace{-0.2cm}
\centering
\tabcolsep=0.3cm
\scalebox{1}{%
\begin{tabular}{l|ccc}
\toprule
\rowcolor{C7!50} Methods          & OBJ-BG & OBJ-ONLY & PB-T50-RS \\ \midrule
PointNet~\cite{qi2017pointnet}           & 73.3   & 79.2     & 68.0      \\
\rowcolor{C7!50} PointNet++~\cite{qi2017pointnet++}      & 82.3   & 84.3     & 77.9      \\
DGCNN~\cite{wang2019dynamic}             & 82.8   & 86.2     & 78.1      \\
\rowcolor{C7!50}  PointCNN~\cite{li2018pointcnn}          & 86.1   & 85.5     & 78.5      \\
SpiderCNN~\cite{xu2018spidercnn}            & 77.1   & 79.5     & 73.7      \\
\rowcolor{C7!50}  BGA-DGCNN~\cite{uy2019revisiting}          & -   & -     & 79.7      \\
BGA-PN++~\cite{uy2019revisiting}             & -  & -   & 80.2      \\
\midrule
Transformer~\cite{yu2022point}     & 79.9  & 80.6    & 77.2     \\
\rowcolor{C7!50} \begin{tabular}[c]{@{}c@{}}  Transformer\\ +OcCo~\cite{wang2021unsupervised} \end{tabular}   & 84.9  & 85.5    & 78.8     \\
Point-BERT~\cite{yu2022point}      & 87.43  & 88.12    & 83.07     \\ 
\rowcolor{C7!50}  Point-MAE~\cite{pang2022masked}          & 90.02   & 88.29    & 85.18   \\
Point-MAE (Rep.)     & 89.36  & 88.68    & 83.83     \\ 
\rowcolor{C7!50} Point-M2AE~\cite{zhang2022point}   & \underline{91.22}  & \underline{88.81}    & \textbf{86.43}     \\
\midrule
\rowcolor{C7!50} \textbf{GS-PT}    & \textbf{91.80} & \textbf{89.44}   &  $\underline{86.09}$   \\ \bottomrule
\end{tabular}%
}
\vspace{-0.7cm}
\label{table:scanobjectnn}
\end{table}

\begin{table}[t]
\caption{\textbf{Few-shot classification on ModelNet40.} Accuracy ($\%$) are listed.
}
\vspace{-0.2cm}
\centering
\tabcolsep=0.07cm

\scalebox{1}{
\begin{tabular}{l|cccc}
\toprule
\rowcolor{C7!50} 
\multicolumn{1}{c|}{\cellcolor{C7!50}}                         & \multicolumn{2}{c|}{\cellcolor{C7!50} 5-way}                         & \multicolumn{2}{c}{\cellcolor{C7!50} 10-way}             \\ \cline{2-5} 
\rowcolor{C7!50} 
\multicolumn{1}{c|}{\multirow{-2}{*}{\cellcolor{C7!50} Method}} & 10-shot             & \multicolumn{1}{c|}{\cellcolor{C7!50} 20-shot} & 10-shot             & 20-shot                                  \\ \midrule

DGCNN~\cite{wang2019dynamic}             & 91.8 ± 3.7          & 93.4 ± 3.2          & 86.3 ± 6.2          & 90.9 ± 5.1                               \\
\rowcolor{C7!50} DGCNN + OcCo~\cite{wang2021unsupervised}                              & 91.9 ± 3.3          & 93.9 ± 3.1          & 86.4 ± 5.4          & 91.3 ± 4.6                               \\ \midrule

\multicolumn{1}{l|}{\cellcolor[HTML]{FFFFFF}Transformer~\cite{yu2022point}}      & 87.8 ± 5.2          & 93.3 ± 4.3          & 84.6 ± 5.5          & 89.4 ± 6.3                               \\
\rowcolor{C7!50} \multicolumn{1}{l|}{Transformer + OcCo~\cite{wang2021unsupervised}}                        & 94.0 ± 3.6          & 95.9 ± 2.3          & 89.4 ± 5.1          & 92.4 ± 4.6                               \\

\multicolumn{1}{l|}{\cellcolor[HTML]{FFFFFF}Point-BERT~\cite{yu2022point}}        & 94.6 ± 3.1          & 96.3 ± 2.7          & 91.0 ± 5.4          & 92.7 ± 5.1                               \\
\rowcolor{C7!50} \multicolumn{1}{l|}{Point-MAE~\cite{pang2022masked}}                        & 96.3 ± 2.5           & 97.8 ± 1.8          & \textbf{92.6 ± 4.1}          & 95.0 ± 3.0                               \\

\multicolumn{1}{l|}{\cellcolor[HTML]{FFFFFF}Point-M2AE~\cite{zhang2022point}}        & \textbf{96.8 ± 1.8}          & 98.3 ± 1.4          & 92.3 ± 4.5          & 95.0 ± 3.0                               \\ \midrule

\rowcolor{C7!50} \multicolumn{1}{l|}{\cellcolor[HTML]{FFFFFF}\textbf{GS-PT}} & \underline{96.5 ± 1.6}    & \textbf{98.7 ± 1.1} & \underline{92.3 ± 4.4}    & \textbf{95.4 ± 3.2}                      \\ \midrule
\end{tabular}
}
\vspace{-0.4cm}
\label{table:Few-shot classification}
\end{table}

\begin{table}[t]
\caption{\textbf{Part segmentation on ShapeNetPart.} `mIoU$_C$' ($\%$) and `mIoU$_I$' ($\%$) respectively represent the average IoU of all component categories and all instances in the dataset. `Rep.' means we reproduce these methods.}
\vspace{-0.2cm}
\centering
\tabcolsep=0.3cm
\scalebox{1}{%
\begin{tabular}{l|cc}
\toprule
\rowcolor{C7!50} Methods         & mIoU$_C$          & mIoU$_I$ \\ \midrule
PointNet~\cite{qi2017pointnet}           & 80.39          & 83.70      \\
\rowcolor{C7!50} PointNet++~\cite{qi2017pointnet++}      & 81.85          & 85.10      \\
DGCNN~\cite{wang2019dynamic}             & 82.33          & 85.20     \\
\midrule
\rowcolor{C7!50} Transformer~\cite{yu2022point}     & 83.42          & 85.10     \\
\begin{tabular}[c]{@{}c@{}}  Transformer\\ +OcCo~\cite{wang2021unsupervised} \end{tabular}   & 83.42          & 85.10     \\
\rowcolor{C7!50} Point-BERT~\cite{yu2022point}      & 84.11          & 85.60    \\ 
Point-MAE~\cite{pang2022masked}          & 84.19          & 86.10   \\
\rowcolor{C7!50} Point-M2AE~\cite{zhang2022point}   & \underline{84.86}    & \textbf{86.51}  \\     
Point-M2AE(Rep.) & 84.75          & 86.35          \\ 
\midrule
\rowcolor{C7!50} \textbf{GS-PT}     & \textbf{85.26} & \underline{86.47}  \\ \bottomrule  
\end{tabular}%
}
\vspace{-0.7cm}
\label{table:Part segmentation}
\end{table}

\begin{table}[t]\small
\caption{\textbf{Classification
results with Linear SVM  on ModelNet40  and ScanObjectNN.} Accuracy ($\%$) are listed. Evaluating Model Performance Through Component Removal in GS-PT.}
\vspace{-0.2cm}
\centering
\tabcolsep=0.07cm
\scalebox{1}{%
\begin{tabular}{c| c  c}
\toprule
\rowcolor{C7!50} \textbf{Alignment Setting}           &  ModelNet40  &  ScanObjectNN      \\  \midrule
Point-M2AE + D                   & 92.30 & 79.52      \\  \midrule
\rowcolor{C7!50}Point-M2AE + P + D        & 92.42 & 80.55       \\  \midrule
    Point-M2AE + I + D        & \underline{92.63} & 80.55       \\  \midrule
\rowcolor{C7!50}Point-M2AE + P + I & 92.50 &  \underline{81.07} \\  \midrule
GS-PT & \textbf{92.67} &  \textbf{82.44}  \\ \bottomrule

\end{tabular}%
}
\label{table:ablation module}
\vspace{-0.2cm}
\end{table}

\begin{table}[t]\small
\caption{\textbf{Linear classification results on ModelNet40 and ScanObjectNN with varying numbers of novel view images.}}
\vspace{-0.2cm}
\centering
\tabcolsep=0.07cm
\scalebox{1}{%
\begin{tabular}{c| c  c}
\toprule
\rowcolor{C7!50} \textbf{Novel View Number (n)}           &  ModelNet40  &  ScanObjectNN      \\  \midrule
1 & \textbf{92.67} & \textbf{82.44} \\ \midrule
\rowcolor{C7!50}2 & 92.34 & 79.17 \\ \midrule
4 & 92.34 &	\underline{81.76} \\ \midrule
\rowcolor{C7!50}6 & 92.42	&79.17 \\ \midrule
8 & \underline{92.54}	& 80.38   \\ \bottomrule

\end{tabular}%
}
\label{table:ablation num views}
\vspace{-0.6cm}
\end{table}

\textbf{Transformed Point Cloud Generation and Novel View Synthesis for Enhanced Triplet Alignment.} 
As illustrated in Fig~\ref{fig:pipeline1}, we adopt 3DGS for SSL in point clouds for the first time and devise a 3DGS branch to formulate an intra-modal and cross-modal correspondence by generating a novel view 2D image and the transformed point cloud. 
Specifically, a U-Net based model~\cite{tang2024lgm} is leveraged to predict 3D Gaussians from multi-view images from our On-the-fly Image Rendering. 
For each object $i$, the U-Net takes four rendered images with corresponding camera pose embeddings as input and predicts a set of 3D Gaussians. The fused 3D Gaussians are obtained by concatenating from these outputs, and then used to extract point clouds $P_{i_{GS}}$ and novel view image $I_{i}$.

\subsection{Aligning Representations of Three Modalities}
\label{sub:Aligning}
With the created triplets of point cloud, novel view image and depth map, 
GS-PT conducts pre-training to align representations of three modalities into the same feature space. Specifically, we train individual feature extractors for each of these modalities and align the point cloud feature with the features of the image and depth map.

\textbf{Intra-modal Contrastive Learning.}
We formulate our intra-modal contrastive learning to enforce geometric invariance between a pair of point clouds.
Given an object $i$,  we predict $\hat{P}_{i}$ through the encoder and decoder, and extract $P_{i_{GS}}$ from Gaussians.
$\hat{P}_{i}$ and $P_{i_{\text{GS}}}$ are considered as a positive pair which represents the spatial information of $i$.
The point cloud feature extractor $f_{\theta_{P}}$ takes the positive point cloud pair as input and outputs point features $\mathbf{z}_{i}$ and $\mathbf{z}_{2i}$, 
\begin{equation}
   \mathbf{z}_{i} = f_{\theta_{P}}(\hat{P}_{i}),
\end{equation}
\begin{equation}
   \mathbf{z}_{2i} = f_{\theta_{P}}(P_{i_{GS}}).
\end{equation}
We perform instance discrimination by pushing closer the distance between a positive point features pair, while pulling away that of negative pairs in a minibatch of examples.
The intra-modal loss function $l(\mathbf{z}_{i},  \mathbf{z}_{2i})$ among the pair $\mathbf{z}_{i}$ and $\mathbf{z}_{2i}$ is computed as:
\begin{equation}
\label{eq:intra-loss}
   l(\mathbf{z}_{i},  \mathbf{z}_{2i}) =-\log \frac{\exp(\frac{\text{s}(\mathbf{z}_{i},  \mathbf{z}_{2i})}{\tau})}{ \sum_{\substack{k=1 \\ k \neq i} }^{2N} \exp(\frac{\text{s}(\mathbf{z}_{i}, \mathbf{z}_{k})}{\tau})},
\end{equation}
where $N$ is the minibatch size, $\tau$ is a temperature parameter and $s(\cdot)$ denotes the cosine similarity function.
The final loss is computed across all positive pairs:
\begin{equation}
    \mathcal{L}_{IM} = \frac{1}{2N} \sum_{k=1}^{N} [l({\mathbf{z}}_{k}, {\mathbf{z}}_{2k}) + l({\mathbf{z}}_{2k}, {\mathbf{z}}_{k})].
\end{equation}

\textbf{Cross-modal Contrastive Learning.}
As illustrated in Fig~\ref{fig:pipeline1}, we embed the rendered novel view image $I_{\text{i}}$ and depth map $D_{\text{i}}$ to a feature space using the feature extractors $f_{\theta_{I}}(\cdot)$ and $f_{\theta_{D}}(\cdot)$,
\begin{equation}
   \mathbf{h}_{i}^{rgb} = f_{\theta_{I}}(I_{{i}}),
\end{equation}
\begin{equation}
   \mathbf{h}_{i}^{depth} = f_{\theta_{D}}(D_{{i}}).
\end{equation}
The point cloud feature is represented as the mean of $\mathbf{z}_{i}$ and   $\mathbf{z}_{2i}$. We aim to maximize the similarity of each pair of modalities corresponding to same object $i$.
Then the contrastive loss of point-image pair is computed as follows, 
\begin{equation}
   l_{c}(\mathbf{\bar{z}_{i}},  \mathbf{h}_{i}^{rgb}) =-\log \frac{\exp(\frac{\text{s}(\mathbf{\bar{z}_{i}},  \mathbf{h}_{i}^{rgb})}{\tau})}{ \sum_{\substack{k=1} }^{N} \exp(\frac{\text{s}(\mathbf{\bar{z}_{i}}, \mathbf{h}_{k}^{rgb})}{\tau})},
\end{equation}
\begin{equation}\small
    \mathcal{L}_{CM}(P,I) = \frac{1}{2N} \sum_{k=1}^{N} [l_{c}({\mathbf{\bar{z}}}_{k}, {\mathbf{h}}_{k}^{rgb}) + l_{c}({\mathbf{h}}_{k}^{rgb}, {\mathbf{\bar{z}}}_{k})],
\end{equation}
where $N$, $\tau$ and $s(\cdot)$ refers to the same parameters as in Eq. \ref{eq:intra-loss}.
The calculation of $\mathcal{L}_{CM}(P,D)$ follows the same principle.

Following~\cite{zhang2022point}, we compute the reconstruction loss $\mathcal{L}_{CD}$ between predicted and ground-truth point cloud coordinates by $l_{2}$ Chamfer Distance.
Finally, we minimize $\mathcal{L}_{total}$ for all intra-modal and cross-modal loss with different coefficients,
\begin{equation}\small
\label{eq:total-loss}
    \mathcal{L}_{total} = \alpha\mathcal{L}_{IM} + \beta\mathcal{L}_{CM}(P,I) + \gamma\mathcal{L}_{CM}(P,D) + \delta\mathcal{L}_{CD},
\end{equation}
where $\alpha$, $\beta$, $\gamma$ and $\delta$ are hyper-parameters.

\section{Experiments}
\label{sec:Experiments}


\subsubsection{Downstream Tasks}
\textbf{3D Object classification.}
As shown in Table~\ref{table:modelnet40}, GS-PT achieves 93.8\% classification accuracy on ModelNet40~\cite{wu20153d}, ranking second only to Point-M2AE~\cite{zhang2022point}, which reported 94.0\% in their original paper.
For a fair comparison, we also reproduce Point-MAE~\cite{pang2022masked} and Point-M2AE using their official codes under the same setups.
As a result, Point-MAE achieves 93.1\%, while Point-M2AE fulfills 93.5\% on ModelNet40.
In the experiments conducted under the same setups, our GS-PT outperforms Point-MAE and Point-M2AE by 0.7\% and 0.3\% in terms of accuracy, respectively. 
For ScanObjectNN~\cite{uy2019revisiting} in Table~\ref{table:scanobjectnn}, Our GS-PT largely improves the baseline by 11.9\%, 8.84\% and 8.89\% for three variants respectively.
For OBJ-BG and OBJ-ONLY, GS-PT outperforms the previous state-of-the-art results achieved by Point-M2AE, indicating a strong generalization capability.

\textbf{Few-shot Learning.}
Following~\cite{yu2022point}, we conduct few-shot learning experiments on ModelNet40, and the results are shown in Table~\ref{table:Few-shot classification}. 
GS-PT exhibits smaller deviations on the four settings, and achieves the optimal performance under ``5-way 20-shot'' and ``10-way 20-shot'', surpassing Point-MAE by +0.9\% and +0.4\%, respectively. 
Under the ``5-way 10-shot'' and ``10-way 10-shot'', GS-PT narrowly trails the best results by -0.3\% in each case, respectively.
These results indicate that GS-PT fulfills the best overall in the few-shot classification, learning more general knowledge for well adapting to new tasks under low-data conditions.

\textbf{Part Segmentation.}
Moreover, we evaluate the representation learning capacity of our GT-PT on ShapeNetPart~\cite{yi2016scalable}.
Table.~\ref{table:Part segmentation} shows the average IoU of all categories and all instances.
We also reproduce Point-M2AE under the same setups for a fair comparison.
As shown in Table.~\ref{table:Part segmentation}, GS-PT improves the baseline by 1.84\% mIoU$_C$ and 1\% mIoU$_I$.
GS-PT achieves the best 85.26\% category mIoU, surpassing the second-best Point-M2AE by +0.4\%. 
In experiments conducted under the same setups, GS-PT outperforms Point-M2AE by +0.12\% instance mIoU and +0.51\% category mIoU.
As illustrated in Fig~\ref{fig:Visualization of part segmentation}, the visualization demonstrates that the segmentation achieved by our GS-PT closely aligns with the ground truth.
These experimental results highlight the learning potential for geometric structures of our GS-PT, attributed to the integration of 3DGS for self-supervised learning.

\textbf{Visualization}
We use t-SNE~\cite{van2008visualizing} to visualize the features extracted by GS-PT, as shown in Fig~\ref{fig:tSNE visualization}. 
These results reveal that GS-PT is capable of generating discriminative features for various categories after pre-training. 
Furthermore, its ability to distinguish categories is greatly enhanced after fine-tuning. 
The results indicate that GS-PT can maintain good performance across various types of datasets, showcasing robust generalization capabilities.

\subsubsection{Ablation Study}
\textbf{Impact of Align Representations.}
As described in Eq.~\ref{eq:total-loss}, our approach aims to train the model by aligning the 3D representation with both the intra-modal and cross-modal representations. 
We investigate whether the performance of our model is affected by the elimination of different modalities through the removal of specific modalities.
We conduct an ablation study for GS-PT by removing one of the modalities at a time and evaluating a linear SVM classifier in both ModelNet40 and ScanObjectNN datasets. 
Results are shown in Table.~\ref{table:ablation module}, which indicates that the best classification performance is achieved when the point clouds and depth images are aligned with 3DGS points and novel-view images.
These results highlight the effectiveness of integrating multiple modalities.

\textbf{Number of Novel View Images.}
We further perform an ablation study to evaluate the contribution of 3DGS novel view images rendering branch by varying the number of rendered views. 
Results in Table.~\ref{table:ablation num views} demonstrate that rendering a single image is also able to benefit multi-modal 3D representation learning to yield better linear SVM classification performance, achieving an accuracy of 92.67\% on the ModelNet40 and 82.44\% on ScanObjectNN. 
This indicates that even a solitary novel view is sufficient to enhance multi-modal 3D representation learning.

\section{Conclusion}
\label{sec:Conclusion}

This paper introduces GS-PT, a unified tri-modal pre-training framework. GS-PT integrates 3D Gaussian Splatting for the first time to pre-train a Transformer backbone using tri-modal alignment objectives, improving its comprehensive understanding of point clouds. 
Experimental results indicate that GS-PT outperforms the off-the-shelf self-supervised learning methods on various downstream tasks.

\newpage
\clearpage
\bibliographystyle{IEEEbib}
\bibliography{refs}

\end{document}


\title{GS-PT: Exploiting 3D Gaussian Splatting for Comprehensive Point Cloud Understanding via Self-supervised Learning
}

\author{
\IEEEauthorblockN{
    Keyi Liu$^{1,\star}$,    
    Yeqi Luo$^{1,\star}$   
    Weidong Yang$^{1}$, 
    Jingyi Xu$^{1}$, 
    Zhijun Li$^{2}$, \IEEEmembership{Fellow, IEEE},  
    Wen-Ming Chen$^{3}$, 
    Ben Fei$^{1}$
    \thanks{$^{\star}$Equal contribution.}
    }
    
    \IEEEauthorblockA{$^{1}$\textit{School of Computer Science, Fudan University}\\$^{2}$\textit{School of Mechanical Engineering, Tongji University}\\$^{3}$\textit{Academy for Engineering and Technology, Fudan University}\\
    {{$\text{23210240242}|\text{23212010018}|\text{jyxu22}|\text{bfei21}$}@m.fudan.edu.cn, wdyang@fudan.edu.cn, zjli@ieee.org, chenwm@fudan.edu.cn}
    }
}  

\maketitle

\section{Related Works}
\label{sec:Related}

\subsubsection{3D Gaussian Splatting}
Recently, 3DGS has gained significant advancements, attributed primarily to its remarkable rendering speed and ability to synthesize realistic scenes from novel perspectives.
Compared with Neural Radiance Fields (NeRF)~\cite{mildenhall2021nerf}, 3DGS introduces a more compact and efficient representation by utilizing Gaussian distributions to model the uncertainty and density of 3D points.
Therefore, 3DGS has been widely used for surface reconstruction~\cite{guedon2024sugar}, dynamic modeling~\cite{yang2024deformable}, large-scene modeling~\cite{lin2024vastgaussian}, scene manipulation~\cite{chen2024gaussianeditor}, 3D generation~\cite{liang2024luciddreamer}, 3D perception~\cite{zhou2024drivinggaussian} and human modeling~\cite{jiang2024hifi4g}.
However, utilizing 3DGS for point cloud self-supervised learning is still an under-explored area.

\subsubsection{Self-supervised Learning in Point Clouds}
Several methodologies~\cite{fei2024curriculumformer,fei2023self} have been developed and examined for self-supervised learning on point clouds. Generally, existing methods can be categorized into contrastive methods and generative methods.

\textbf{Generative Methods} employ an encoder-decoder architecture to learn representations from point cloud via self-reconstruction~\cite{wang2021unsupervised,yu2022point,pang2022masked,zhang2022point}.
Inspired by the success of the masked auto-encoder (MAE) in 2D computer vision, Pang et al.~\cite{pang2022masked} proposed Point-MAE for 3D point clouds. 
Point-M2AE~\cite{zhang2022point} advances upon Point-MAE by addressing the limitations related to encoding single-resolution point clouds and neglection of local-global relations in 3D shapes.
Through skip connections between encoder and decoder stages, Point-M2AE enhances fine-grained information during up-sampling, promoting local-to-global reconstruction and capturing the relationship between local structure and global shape.

\textbf{Contrastive Methods} learn discriminative features by training network to distinguish between positive and negative samples~\cite{xie2020pointcontrast,afham2022crosspoint,huang2021spatio}.
CrossPoint~\cite{afham2022crosspoint} is a cross-modal contrastive learning method, which introduces a contrastive loss between the rendered 2D image feature and the point cloud feature.
However, existing works only consider point cloud and RGB information. Our aim is to leverage the correlation across 3D point clouds, 2D images, and depth maps via contrastive learning to learn comprehensive 3D representations.

\section{Experimental Setups}
We conduct the following experiments with our GS-PT.
First, we pre-train our model on ShapeNet~\cite{chang2015shapenet}. 
Next, we evaluat the pre-trained model on various downstream tasks, including 3D object classification, few-shot learning, and part segmentation. Additionally, we perform ablation studies by training the model with different learning objective combinations to assess their contributions.

\subsection{Implementation Details}

\textbf{Self-supervised Pre-training.}
We pre-train GS-PT on 2 NVIDIA A100 GPUs. 
For the 3D input, we uniformly sample $N_{p}$ = 2,048 points and resized the rendered images into $256 \times 256$.
We pre-train the network for $20$ epochs by fine-tuning the pre-training checkpoint from Point-M2AE~\cite{zhang2022point}, with a batch size $8$.
The AdamW~\cite{diederik2014adam} optimizer is employed, with an initial learning rate set to $10^{-4}$  and a weight decay of $5 \times 10^{-2}$.
Cosine annealing is employed as the learning rate scheduler.
For multi-modalities alignment, we deploy DGCNN~\cite{wang2019dynamic} as the point cloud feature extractor $f_{\theta_{P}}(\cdot)$, and two ResNet-50~\cite{he2016deep} networks as the image feature extractor $f_{\theta_{I}}(\cdot)$ and depth feature extractor $f_{\theta_{D}}(\cdot)$s, respectively.
For the setting of $\mathcal{L}_{total}$, $\alpha$, $\beta$, and $\gamma$ is set to be  $10^{-3}$, while $\delta$ is set to be 1 to balance the contributions of various loss components during pre-training.
$\mathcal{L}_{CD}$ between reconstructed and ground-truth point cloud coordinates is computed as 
\begin{equation}\small
\label{eq:CD loss}
\mathcal{L}_{CD}(\hat{P_{i}}, P_{i}) = \frac{1}{|\hat{P_{i}}|} \sum_{p \in \hat{P_{i}}} \min_{q \in P_{i}} \| p - q \|^2 + \frac{1}{|P_{i}|} \sum_{q \in P_{i}} \min_{p \in \hat{P_{i}}} \| q - p \|^2.  
\end{equation}

All downstream tasks are conducted on a single NVIDIA A100 GPU.

\textbf{3D Object classification.} For downstream classification tasks, we uniformly sample 1,024 points from each object in ModelNet40~\cite{wu20153d} and 2,048 points from ScanObjectNN~\cite{uy2019revisiting}, respectively. 
We use the encoded features of point tokens as input to the classification head.
We apply max pooling and average pooling to these features, then sum the results which contain the global representations.
The classification head is constructed with MLP layers, each of which is adopted with batch normalization, ReLU activation, and dropout operation except the last layer.  
We fine-tune the network for 300 epochs with a batch size of 32.
The AdamW optimizer~\cite{diederik2014adam} is employed, with an initial learning rate set at $5 \times 10^{-4}$ and a weight decay of $5 \times 10^{-2}$.
For testings on ModelNet40, the voting method follows~\cite{liu2019relation}.

\textbf{Few-shot Learning.}
Following previous works~\cite{yu2022point,afham2022crosspoint,wang2021unsupervised}, we conduct few-shot learning experiments on ModelNet40 with a typical setting of \textit{K-way N-shot}.
Here, $K$ classes are randomly sampled from the dataset, while $N$ objects are randomly sampled for each class to train the network.
We adopt $K\in\{5,10\}$ and $N \in \{10,20\}$, creating 4 group settings, and sample additional 20 objects per class for testing.
To improve the reliability of the results, we conduct 10 independent experiments under each setting and report the average accuracy and standard deviation.
The fine-tuning settings are the same as 3D object classification experiments.

\textbf{Part Segmentation.}
For each 3D object, we sample 2,048 points.
For each stage of the hierarchical encoder, the output features of different scales are fed into feature propagation block~\cite{qi2017pointnet++}, thus up-sampled to point features of corresponding point coordinates.
These features are concatenated together and then subjected to average pooling and max pooling separately.
We combine the sum of pooling results with the class feature and point feature through concatenation.
The segmentation head which is constructed with stacked 1D convolutional layers takes the concatenation result as input and predicts label of each point.
We fine-tune the network for 300 epochs with a batch size of 16, with an initial learning rate set at $2 \times 10^{-4}$ and a weight decay of 0.1.
\subsection{Datasets}
\subsubsection{Pre-training Dataset}
We pre-train our GS-PT on ShapeNet~\cite{chang2015shapenet} dataset, which contains more than 50,000 synthetic 3D shapes of 55 categories, associated with metadata that describes the semantic information of each CAD model.

\subsubsection{Downstream Datasets}

\textbf{3D Object classification.}
We fine-tune our GS-PT on ModelNet40~\cite{wu20153d} and ScanObjectNN~\cite{uy2019revisiting}. ModelNet40 consists of 12,311 clean 3D CAD models from 40 categories, with a split of 9,843 models for training and 2,468 models reserved for testing. ScanObjectiNN is a more realistic and challenging point cloud classification dataset, encompassing 15 categories and approximately 15,000 samples of actual scanned objects. During the experimentation, we evaluate the model's performance on three variants of ScanObjectNN: OBJ-BG, OBJ-ONLY, and PB-T50-RS.

\textbf{Part Segmentation.}
we evaluate the representation learning capacity of our GT-PT on ShapeNetPart~\cite{yi2016scalable} dataset. ShapeNetPart consists of 16,881 3D models from 16 categories, with 50-part segmentation annotations.

\bibliographystyle{IEEEbib}
\bibliography{refs}